\documentclass[runningheads]{llncs}
\usepackage{graphicx}
\usepackage{amsmath}
\usepackage{booktabs}
\usepackage{amssymb}
\usepackage{comment}
\usepackage{todonotes}
\usepackage{hyperref}
\usepackage{multirow}
\usepackage[subtle]{savetrees}
\usepackage{bm}
\begin{document}
\title{Uncertainty quantification in medical image segmentation with normalizing flows}
%\thanks{Supported by organization x.
\titlerunning{cFlow Net}
%\author{First Author \inst{1}
%\orcidID{} 
%Second Author \and
%Third Author \and
%Fourth Author}

\author{Raghavendra Selvan\inst{1}
\and
Frederik Faye \inst{2} \and
Jon Middleton \inst{2} \and
Akshay Pai \inst{1,2}}
\authorrunning{R. Selvan et al.}
%\authorrunning{First Author et al.}
%\institute{Institute 1 \and
%Institute 2 \\
%\email{mail@email}}
\institute{Department of Computer Science, University of Copenhagen, Denmark \and
Cerebriu A/S, Copenhagen, Denmark \\
\email{raghav@di.ku.dk}
}
\def \Em{{\mathbb{E}}}
\def \Rm{{\mathbb{R}}}

\def \Im{{\mathbb{I}}}
\def \abf{{\mathbf a}}
\def \Abf{{\mathbf A}}
\def \bbf{{\mathbf b}}
\def \Bbf{{\mathbf B}}
\def \cbf{{\mathbf c}}
\def \Cbf{{\mathbf C}}
\def \dbf{{\mathbf d}}
\def \Dbf{{\mathbf D}}
\def \ebf{{\mathbf e}}
\def \Ebf{{\mathbf E}}
\def \fbf{{\mathbf f}}
\def \Fbf{{\mathbf F}}
\def \gbf{{\mathbf g}}
\def \Gbf{{\mathbf G}}
\def \hbf{{\mathbf h}}
\def \Hbf{{\mathbf H}}
\def \ibf{{\mathbf i}}
\def \Ibf{{\mathbf I}}
\def \jbf{{\mathbf j}}
\def \Jbf{{\mathbf J}}
\def \kbf{{\mathbf k}}
\def \Kbf{{\mathbf K}}
\def \lbf{{\mathbf l}}
\def \Lbf{{\mathbf L}}
\def \mbf{{\mathbf m}}
\def \Mbf{{\mathbf M}}
\def \nbf{{\mathbf n}}
\def \Nbf{{\mathbf N}}
\def \obf{{\mathbf o}}
\def \Obf{{\mathbf O}}
\def \pbf{{\mathbf p}}
\def \Pbf{{\mathbf P}}
\def \qbf{{\mathbf q}}
\def \Qbf{{\mathbf Q}}
\def \rbf{{\mathbf r}}
\def \Rbf{{\mathbf R}}
\def \sbf{{\mathbf s}}
\def \Sbf{{\mathbf S}}
\def \tbf{{\mathbf t}}
\def \Tbf{{\mathbf T}}
\def \ubf{{\mathbf u}}
\def \Ubf{{\mathbf U}}
\def \vbf{{\mathbf v}}
\def \Vbf{{\mathbf V}}
\def \wbf{{\mathbf w}}
\def \Wbf{{\mathbf W}}
\def \xbf{{\mathbf x}}
\def \Xbf{{\mathbf X}}
\def \ybf{{\mathbf y}}
\def \Ybf{{\mathbf Y}}
\def \zbf{{\mathbf z}}
\def \Zbf{{\mathbf Z}}
\def \Hbf{{\mathbf H}}
\def \0bf{{\mathbf 0}}

\def \Emean{\mathbb{E}}
\def \Rm{\mathbb{R}}

\def \Sibf{{\mathbf \Sigma}}
\def \xbbf{\mathbf{\bar{x}}}
\def \etr{\mbox{etr}}
\def \tr{\mbox{tr}}
\def \Tr{\mbox{Tr}}
\def \Cov{\mbox{Cov}}
\def \cost{\mbox{cost}}
\def \diag{\mbox{diag}}
\def \Lambf{{\mathbf{\Lambda}}}
\def \Gambf{{\mathbf{\Gamma}}}
\def \Sigbf{{\mathbf \Sigma}}
\newcommand{\rhobf}{\ensuremath{\boldsymbol{\rho}}}
\newcommand{\lambf}{\ensuremath{\boldsymbol{\lambda}}}
\newcommand{\nubf}{\ensuremath{\boldsymbol{\nu}}}
\def \alphabf{{\boldsymbol{\alpha}}}
\def \sigmabf{{\boldsymbol{\sigma}}}
\def \mubf{{\boldsymbol{\mu}}}

\def \Ncal{{\mathcal N}}
\def \Pcal{{\mathcal P}}
\def \Fcal{{\mathcal F}}
\def \Ecal{{\mathcal E}}
\def \Scal{{\mathcal S}}
\def \Qcal{{\mathcal Q}}
\def \Bcal{{\mathcal B}}
\def \Lbcal{{\mathcal Lb}}
\def \Gcal{{\mathcal G}}
\def \Lcal{{\mathcal L}}
\def \KLD{\text{KLD}}
\def \KL{\text{KL}}

%\\
\maketitle              % typeset the header of the contribution
\begin{abstract}
Medical image segmentation is inherently an ambiguous task due to factors such as partial volumes and variations in anatomical definitions. While in most cases the segmentation uncertainty is around the border of structures of interest, there can also be considerable inter-rater differences. The class of conditional variational autoencoders (cVAE) offers a principled approach to inferring distributions over plausible segmentations that are conditioned on input images. Segmentation uncertainty estimated from samples of such distributions can be more informative than using pixel level probability scores. In this work, we propose a novel conditional generative model that is based on  conditional Normalizing Flow (cFlow). The basic idea is to increase the expressivity of the cVAE by introducing a cFlow transformation step after the encoder. This yields improved approximations of the latent posterior distribution, allowing the model to capture richer segmentation variations. With this we show that the quality and diversity of samples obtained from our conditional generative model is enhanced. Performance of our model, which we call \emph{cFlow Net}, is evaluated on two medical imaging datasets demonstrating substantial improvements in both qualitative and quantitative measures when compared to a recent cVAE based model.
		%\footnote{Source code is available here: \url{https://github.com/raghavian/cFlow}}
\end{abstract}
\section{Introduction}

%Supervised medical image segmentation requires manual annotations that are formulated as a consensus from multiple raters to account for inter-rater variations. While the current class of segmentation methods rely on definitive annotations from consensus, such inter-rater variations may be used to model segmentation uncertainties which is different from epistemic and aleoteric uncertainties. 

Medical image segmentation is inherently an ambiguous task and segmentation methods capable of quantifying uncertainty by inferring distributions over segmentations are therefore of substantial interest to the medical imaging community~\cite{wilson1988image,kendall2017uncertainties,jensen2019improving}. Estimating uncertainty from distributions over segmentations is closer to the clinical settings, than obtaining pixel-wise uncertainty estimates, where \emph{whenever feasible} multiple expert opinions are used to ascertain downstream clinical decisions. Such consensus based decisions not only account for the aleatoric (inherent) and epistemic (modeling) uncertainties but also explain the inter-rater variability that is largely inevitable in medical image segmentation.

Remarkable strides in supervised medical image segmentation have been made with deep learning methods~\cite{ronneberger2015u,cciccek20163d,zhou2019review}. These methods, however, provide point estimates of segmentations -- meaning a single segmentation mask per image -- which limits our ability to quantify the uncertainty of said segmentations. 

Bayesian deep learning methods offer a natural setting to infer distributions over segmentations. This has been explored to some extent for medical image segmentation in the spirit of Monte Carlo estimation where multiple hypotheses are explored by predicting segmentation masks with different dropout rates~\cite{gal2016dropout} or with an ensemble of models~\cite{rupprecht2017learning}. These methods can output a fixed number of samples with pixel level probability scores which can be a limitation.

Conditional variational autoencoders (cVAE)~\cite{sohn2015learning} belong to the class of conditional generative models. cVAEs can be used to obtain an unlimited number of predictions by sampling from a latent space conditioned on the input images. This model was adapted for medical image segmentation as the probabilistic U-Net (Prob. U-Net)~\cite{kohl2018probabilistic} demonstrating the possibility of generating large number of plausible segmentations. The Prob. U-Net model fuses an additional channel obtained from the latent space to the final layer (at the highest resolution) of U-Net to obtain a variety of albeit less diverse and blurry segmentations when compared to the raters~\cite{baumgartner2019phiseg}. Quite recently, two models have sought to improve upon the Prob. U-Net~\cite{baumgartner2019phiseg,kohl2019hierarchical}. Both these methods hypothesize that the blurriness and lack of diversity observed in samples obtained from Prob. U-Net is caused by the use of a single latent variable at the highest resolution. They propose using latent variables in a hierarchical fashion operating at different resolutions to make the model more expressive and demonstrate this to be helpful.

\begin{figure}[t]
    \centering
    \includegraphics[width=0.7\textwidth]{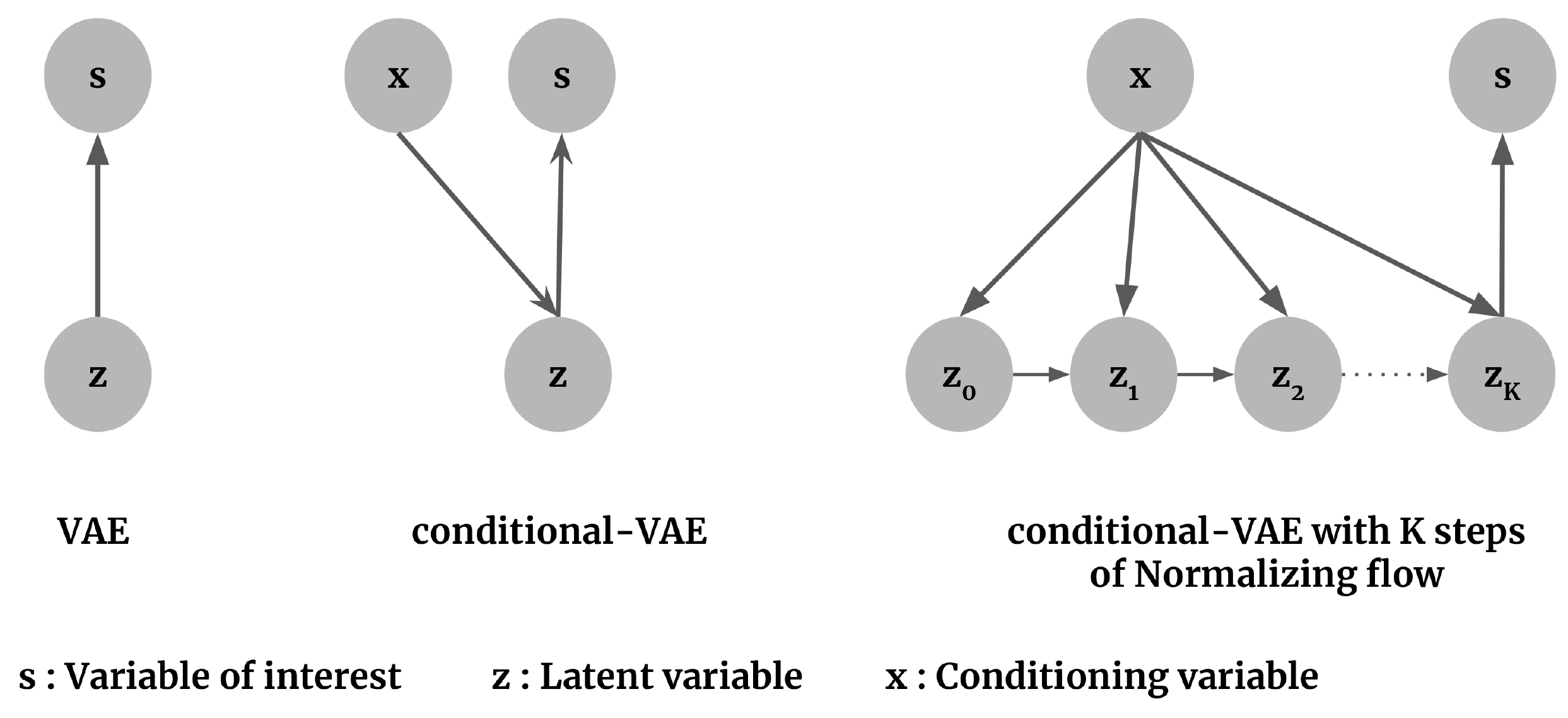}
		\caption{Graphical model view of VAE and variations to it including the proposed cFlow Net (right)}
    \label{fig:graphicalModel}
    \vspace{-0.5cm}
\end{figure}

In this work, we focus on obtaining expressive latent representations that can yield diverse segmentations within the cVAE setting. While we agree with~\cite{baumgartner2019phiseg,kohl2019hierarchical} that Prob. U-Net suffers from using the latent representation at a single resolution, we argue that it can be alleviated by using a more expressive latent posterior distribution instead of using multiple latent variables in a hierarchical setting. This arises from the fact that all cVAE type models, including Prob. U-Net, use an axis aligned Gaussian as the latent distribution which can be limiting when approximating a complex latent posterior distribution~\cite{rezende2015variational,kingma2016improved}. We propose to improve the approximation of the latent posterior distribution with conditional normalizing flows (cFlow) which can yield arbitrarily complex distributions starting from simple ones. We demonstrate that these complex distributions operating at a single resolution are  able to capture richer diversity of realistic segmentations. We propose a novel conditional normalizing flow model -- cFlow Net -- and demonstrate the use of two types of normalizing flow transformations: Planar flows~\cite{rezende2015variational} and Generative Flows~\cite{kingma2018glow}. We evaluate the method on two medical imaging datasets: LIDC-IDRI~\cite{armato2004lung} for detecting lesions in lungs from chest CT and for detecting retina blood vessels from on a new Retinal Vessel dataset created from three older datasets~\cite{staal2004ridge,hoover2000locating,owen2011retinal}.
%The Retinal Vessel dataset was created from three existing datasets with two annotators:  DRIVE~\cite{staal2004ridge}, STARE~\cite{hoover2000locating} and CHASE~\cite{owen2011retinal}. 
We compare the performance of our model with deterministic U-Net~\cite{ronneberger2015u} and Prob. U-Net demonstrating significant improvements on both quantitative (generalized energy distance and dice) and qualitative measures.

\section{Background \& Problem Formulation}
\label{sec:flow}

%\subsection{Conditional Variational Autoencoders}
%\subsection{Normalizing Flows}

Image segmentation tasks can be formulated in a conditional generative model setting with the objective of estimating the conditional distribution $p(\sbf|\xbf)$, where $\xbf \in \Rm^{H\times W \times C}$  and $\sbf \in \{0,1\}^{H \times W} $ are the input images and corresponding binary segmentations, respectively, of dimensions $H,W$ with $C$ channels. This has been approached using the conditional VAE formulation where the conditional distribution $p(\sbf|\xbf)$ is approximated by introducing dependency on a $d$-dimensional latent variable $\zbf \in \Rm^d$~\cite{sohn2015learning,kohl2018probabilistic}, as shown in Figure~\ref{fig:graphicalModel} (center). 

The cVAE objective minimizes the KL divergence between the true latent posterior distribution $p(\zbf|\sbf,\xbf)$ and its variational approximation $q(\zbf|\sbf,\xbf)$ resulting in an objective of the form~\cite{sohn2015learning}: 
\begin{equation}
    \mathcal{L}_{\text{cVAE}} =  -\Em_{q_{\phi}(\zbf|\sbf,\xbf)} \big [ \log p_\theta(\sbf|\zbf,\xbf)\big] + \text{KL} \big[q_{\phi}(\zbf|\sbf,\xbf) || p_\psi(\zbf|\xbf)\big]
    \label{eq:cVae}
\end{equation}
The first term is the expected conditional log-likelihood (CLL) under the variational distribution $q_\phi(\zbf|\sbf,\xbf)$ and the second term can be seen as the regularization forcing the posterior distribution to match the conditional prior distribution $p_\psi(\zbf|\xbf)$. In cVAE, the posterior density is modeled as a diagonal Gaussian density for tractability reasons: $q_{\phi}(\zbf|\sbf,\xbf) = N(\zbf;\mathbf{\mubf}_{\phi},{\sigmabf}^2_{\phi})$. The mean $\mubf_\phi$ and variance $\sigmabf^2_\phi$ are predicted using an encoder network parameterized by $\phi$. The decoder and prior networks are parameterized by $\theta$ and $\psi$, respectively.

Normalizing flows can be used to transform simple base distributions into complex ones using a sequence of bijective transformations (the flow chain) with easy to compute Jacobians~\cite{rezende2015variational,papamakarios2019normalizing}. They basically extend the change of variable rule to transform a base distribution into a target distribution in $K$ successive steps. 
%Under reasonable conditions on the target distributions, $p(\zbf_K)$, 
Normalizing flows can transform a simple base distribution $p(\zbf_0)$ into an arbitrarily complex target distribution, $p(\zbf_K)$, 
by composing complex flow transformations with simpler flow steps~\cite{papamakarios2019normalizing}. 
%Further, due to their reversible nature enabled by the flow chain there is an increased interest in designing invertible neural networks (INN). INNs have seen applications in solving inverse problems and in developing very deep neural networks without a large increase in memory footprint as intermediate feature maps for backward computation can be analytically computed~\cite{lyton2018,invert2019}.

Consider one such bijective transformation $T$ composed of $K$ steps:
\begin{equation}
    T = T_K \circ T_{K-1} \circ \cdots T_{1}.
    \label{eq:flowTrans}
\end{equation}
Forward evaluation of this flow chain, transforming $\zbf_0 \rightarrow \zbf_K$, can be written as: 
\begin{equation}
   \zbf_k = T_k (\zbf_{k-1}) \quad \text{for }k=1\dots K
\end{equation} % FF: Maybe instead "\text{for }k=1..K"? When I see ":" in a mathematical context, I think of it as e.g. f:X->Y etc.
where $\zbf_0$ is distributed according to the base distribution $p(\zbf_0)$. \\
Reverse evaluation of the flow chain, transforming $\zbf_K \rightarrow \zbf_0$, can be written as: 
\begin{equation}
    \zbf_{k-1} = T_k^{-1} (\zbf_{k}) \quad \text{for } k=K \dots 1.
\end{equation}
The transformed distribution, $p(\zbf_K)$, is obtained from the base distribution, $p(\zbf_0)$, adjusted by the inverse absolute Jacobian determinant of the flow transformation. For a single flow step $k$: 
\begin{equation}
    p(\zbf_k) = p(\zbf_{k-1})\Big |\frac{\partial T_k(\zbf_{k-1})}{\partial \zbf_{k-1}}\Big |^{-1} = p(\zbf_{k-1})\Big |J_{T_k}(\zbf_{k-1})\Big |^{-1}
\end{equation}
where $J_{T_k}(\zbf_{k-1})$ denotes the Jacobian determinant.
The complete transformation using the full flow chain in log domain is given by
\begin{align} % FF: Shouldn't it be "\log p(\zbf_K) = "?
    \log p(\zbf_K) = \log p(\zbf_0) + \log \Big{|}  \prod_{k=1}^K J_{T_k}^{-1}(\zbf_{k-1})\Big{|} %\nonumber \\
    = \log p(\zbf_0) - \sum_{k=1}^K \log \Big{|}  J_{T_k}(\zbf_{k-1})\Big{|},
    \label{eq:flowDensity}
\end{align}
where the last equality follows from $\log \left| J_{T_k}^{-1} \right| = \log \left| J_{T_k}\right|^{-1}  = - \log \left| J_{T_k} \right|$.
%is due to a property of Jacobians of bijective functions. %% %FF: "where the last equality follows from $\log \left| J_{T_k}^{-1} \right| = \log \left| J_{T_k}\right|^{-1}  = - \log \left| J_{T_k} \right|$."
%\todo{check!}
%We use planar flows
%two types of normalizing flows in this work: Planar flows~\cite{rezende2015} and Generative flows~\cite{kingma2018} which are briefly presented next.

%\subsubsection{Generative flow (Glow):} A single step in the Glow model introduced in~\cite{kingma2018} comprises a data dependent initialisation layer (ActNorm, short for Activation Normalization), invertible $1\times 1$ convolution and an affine coupling layer~\cite{nice2014}. The forward evaluations of these steps and the corresponding Jacobian determinants are 
%\section{Methods}

%\subsection{cFlow Net: Improving cVAE with Normalizing Flows}

\vspace{-0.2cm}
\section{Methods}
\label{sec:cFlowNet}
\vspace{-0.2cm}
%The constraint on the posterior distribution to be an axis-aligned Gaussian in cVAE can be highly restrictive and in many cases cannot yield good approximations of the true posterior distribution $p(\zbf|\sbf,\xbf)$. 
When using cVAE-like models for medical image segmentation tasks, it is assumed that the diversity of segmentations is captured with the latent posterior distribution. However, using a simple distribution such as an axis-aligned Gaussian to approximate the latent posterior distribution can be too restrictive and might not be sufficiently expressive to capture richer variations. This is noticeable in the Prob. U-Net model~\cite{kohl2018probabilistic} where the segmentations are blurry and lack diversity~\cite{baumgartner2019phiseg,kohl2019hierarchical}. It is in this context that normalizing flows can be used to improve the flexibility of the approximate posterior density to capture a richer diversity of high quality segmentations.
% FF: This is really nice and clear, and motivates this work well!

If we denote the approximate posterior density output by the encoder network as the base distribution, $q(\zbf_0|\sbf,\xbf)$, using the latent variable $\zbf_0$, then using the idea of normalizing flows in Section~\ref{sec:flow} can yield more expressive posterior densities. If the base distribution is transformed using a flow chain of $K$ steps according to Eq.~\eqref{eq:flowTrans}, then the transformed distribution after $K$ steps with $\zbf = \zbf_K$  can be written using Eq.~\eqref{eq:flowDensity} as:
\begin{equation} % FF: The ";x" in "\zbf_{k-1};\xbf" is not completely clear to me. Is it a conditional? If so, I think using "|" would be more clear.
     \log q(\zbf|\sbf,\xbf) = 
     \log q(\zbf_K|\sbf,\xbf) =  \log q(\zbf_0|\sbf,\xbf) - \sum_{k=1}^K  \log \Big{|} J_{T_k}(\zbf_{k-1}|\xbf)\Big{|}.
     \label{eq:flow}
\end{equation}
It can be shown that the modified objective for the conditional flow-based model becomes (see Section~\ref{sec:objective} in the supplementary material):
\begin{align}
    \mathcal{L}_{\text{cFlow}} 
& = -\Em_{q_{\phi}(\zbf_0|\sbf,\xbf)} \Big{[} \log p_\theta(\sbf|\zbf_K,\xbf)\Big]  \nonumber \\
     &+ \KL \Big[q_{\phi}(\zbf_0|\sbf,\xbf) || p_\psi(\zbf_K|\xbf)\Big] - \Em_{q_{\phi}(\zbf_0|\sbf,\xbf)} \Big{[} \sum_{k=1}^K  \log \Big{|} J_{T_k}(\zbf_{k-1}|\xbf)\Big{|}\Big].
     \label{eq:cFlow}
\end{align}
Note that the expectation is with respect to the \emph{base} distribution of the normalizing flow $q_\phi(\zbf_0|\sbf,\xbf)$. The KL divergence is similar to the term for cVAE in Eq.~\ref{eq:cVae} except for an additional term due to the log determinant of the Jacobian terms in Eq.~\eqref{eq:flow}.

%where we have used Eq.~\eqref{eq:planar} and Eq.~\eqref{eq:flow} in adapting the cVAE objective in Eq.~\eqref{eq:cVae} for the cFlow Net model, similar to the formulations in~\cite{rezende2015variational,van2018sylvester} for the VAE setting. The complete derivation is provided in the supplementary material.

\begin{figure}[t]
    \centering
    \includegraphics[width=0.95\textwidth]{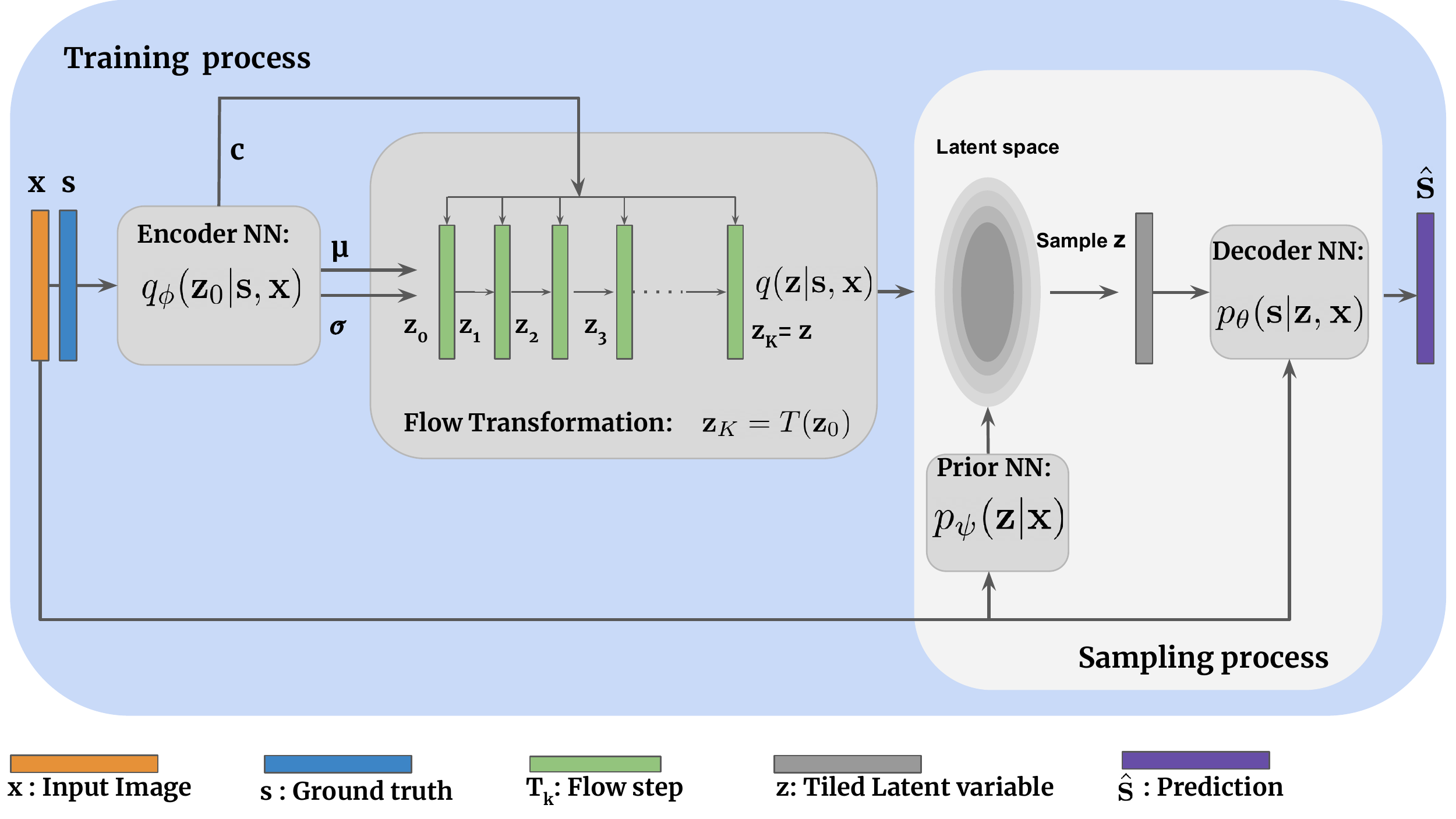}
    \caption{Proposed cFlow Net model. (Training) The training process takes the reference segmentations $\sbf$ and the image data $\xbf$ as input to the encoder, which predicts the mean $\mubf$ and standard deviation $\sigmabf$ of the base distribution along with the context vector $\cbf$ for the flow transformation. The flow transformation block transforms the base distribution, $q_\phi(\zbf_0|\sbf,\xbf)$ to an approximation of the target posterior distribution $q(\zbf|\sbf,\xbf)$ in $K$ steps. The latent space is jointly learned by minimizing the KL divergence between the transformed posterior distribution $q(\zbf|\sbf,\xbf)$ and the conditional prior $p_\psi(\zbf|\xbf)$. (Sampling) The sampling process involves obtaining samples from the conditional prior which is used with the input image together to be decoded in the decoder $p_\theta(\sbf|\zbf,\xbf)$ to obtain the segmentation $\hat\sbf$. After training the model, only the sampling part of the network is used for inference.} 
    \label{fig:cFlowNet}
    \vspace{-0.3cm}
\end{figure}
    \vspace{-0.3cm}

\subsubsection{Planar Flows:}
\vspace{-0.2cm}
In this work we use planar flows introduced in~\cite{rezende2015variational} modified to be conditioned on the input image $\xbf$ with each step of the flow:
\begin{align}
    \ubf_k,\wbf_k, b_k &= f_k(\xbf) \\
    T(\zbf_k|\xbf) &= \zbf_{k-1} + \ubf_k h(\wbf_k^T \zbf_{k-1} + b_k)
    \label{eq:planarForward}
\end{align}
where $\{\ubf_k,\wbf_k \in \Rm^L, b_k \in \Rm \}$ are learnable parameters predicted by a conditioning neural network $f_k(\cdot)$ similar to the conditioning network used in~\cite{lu2019structured}, $L$ is the dimensionality of the latent space and $h(\cdot)$ is an element-wise non-linearity such as \emph{tanh} with derivative $h^\prime(\cdot)$. % FF: I would omit the brackets {}.
The Jacobian determinant for the planar flow step $T_{k}$ is given by
\begin{equation}
    \Big |J_{T_k}(\zbf_{k-1}|\xbf)\Big| = \Big| 1+\ubf_k^T \psi_k(\zbf_{k-1}) \Big| \quad \text{where }  \psi_k(\zbf_{k-1}) = {h}^\prime(\wbf_k^T \zbf_{k-1} +b_k)\wbf_k.
    \label{eq:planar}
\end{equation}
%Notice now the flow steps are dependent on the input $\xbf$ indicated as the conditioning variable in Eq.~\eqref{eq:planarForward} and~\eqref{eq:planar}. 
The conditioning on the flow chain is introduced through the context vector $\cbf$ which is dependent on $\xbf$. The context vector $\cbf \in \Rm^H$ of dimension $H$ is also predicted by the encoder network. The proposed cFlow Net model is visualized in Figure~\ref{fig:cFlowNet}.  

Note that at inference, to sample multiple segmentations only the \emph{Sampling Process} part of the model is used. Given an image $\xbf$, the prior network can be used to obtain multiple latent variable samples $\zbf$  which are then decoded by the decoder network to output multiple segmentations for the input image.
%This dependence on the flow steps on the input via the context vector $\cbf$ is how we 
%We modify the encoder neural network to output a context vector $\cbf$, along with the mean and variance for the base distribution $q(\zbf_0|\sbf,\xbf)$ which is then used to parameterise the flow steps 
\section{Experiments \& Results}
%\vspace{-0.25cm}
\subsection{Data}
%\vspace{-0.25cm}

All experiments are performed on two publicly available datasets. Both datasets comprise labels from at least two raters used to quantify the performance of all models. We use a training-validation-test split of 60:20:20 for both datasets. \\
{\bf LIDC-IDRI dataset:} The LIDC-IDRI dataset consists of $1018$ thoracic CT scans with four raters annotating the lesions in them~\cite{armato2004lung}. We use patches of size $128\times 128$ centered on lesions similar to the procedures followed in~\cite{kohl2018probabilistic,baumgartner2019phiseg} to obtain $15,096$ patches in total. The preprocessed data is obtained from~\cite{lidc}. 
\\
{\bf Retinal Vessel dataset:} As a secondary dataset we create a new dataset derived from three older retinal vessel segmentation datasets: DRIVE~\cite{staal2004ridge}, STARE~\cite{hoover2000locating} and CHASE~\cite{owen2011retinal}. Each of these datasets has a subset of images with labels from two raters. We collected images with two raters from these three datasets, extracted retinal masks when there were none and resized them such that all images are of height $512$ px.  This yields $68$ images of which $20$ are of size $620\times 512$ px and the remaining $48$ are $512\times 512$ px. All images have vessel annotations from two raters. (Figure~\ref{sec:retina} in the supplementary material).

\vspace{-0.4cm}
\subsection{Experiments and Results}
\vspace{-0.2cm}

The proposed cFlow Net model is compared with the probabilistic U-Net~\cite{kohl2018probabilistic}, and additionally with the deterministic U-Net~\cite{ronneberger2015u} for the single rater setting. Other than the cFlow Net model described in Section~\ref{sec:cFlowNet} with planar flows~\cite{rezende2015variational}, we additionally report the cFlow model with conditional generative flow model which uses the Glow transformation steps~\cite{kingma2018glow,lu2019structured} (Section~\ref{sec:glow} in the supplementary material).

Performance of the models in the multiple annotator setting is evaluated based on the generalized energy distance ($d^2_\text{GED}$) which captures the diversity of samples obtained from the generative models when compared to the annotators. It is given by
\begin{equation}
    d^2_\text{GED}(P_{R},P_{M}) = 2 \Em\Big[ d(\sbf,\hat \sbf)\Big] - \Em\Big[ d(\sbf,\sbf^\prime)\Big] - \Em\Big[ d(\hat\sbf,\hat\sbf^\prime)\Big],
    \label{eq:ged}
\end{equation}
where $\sbf, \sbf^\prime $ are samples from the ground truth distribution, $P_R$, comprising different raters, $\hat\sbf, \hat\sbf^\prime $ are samples from the generative distribution, $P_M$, learned by the model and $d(\cdot)$ is 1-IoU ( intersection-over-union) measure. Additionally, we report the negative conditional log likelihood (-CLL $= -\log p (\sbf|\xbf))$ approximated with $128$ samples (Section~\ref{sec:cll} in the supplementary material) and the dice accuracy for the single rater settings.

\begin{table}[t]
\scriptsize
\centering
		\caption{Performance comparison of all models. Higher is better for Dice and lower is better for -CLL and $d^2_\text{GED}$. Significant differences are shown in bold.}
\label{tab:results}
\centering
\begin{tabular}{@{}lc@{\hskip 0.15cm}c@{\hskip 0.15cm}c@{\hskip 0.15cm}c@{\hskip 0.15cm}c@{\hskip 0.15cm}c@{\hskip 0.15cm}c@{\hskip 0.15cm}c@{\hskip 0.15cm}c@{\hskip 0.15cm}c@{\hskip 0.15cm}c@{\hskip 0.15cm}c@{}}
\toprule
\multirow{3}{*}{\textbf{Models}} & &\multicolumn{5}{c}{\textbf{LIDC Dataset}}                                          & \multicolumn{5}{c}{\textbf{Retina Dataset}}                                        \\
\cmidrule(lr){2-6} \cmidrule(lr){7-12}
                        & \multicolumn{2}{c}{All Raters} & \multicolumn{3}{c}{Single Rater}& & \multicolumn{2}{c}{All Raters} & \multicolumn{3}{c}{Single Rater} \\
                        \cmidrule(lr){2-3} \cmidrule(lr){4-6} \cmidrule(lr){8-9} \cmidrule(lr){10-12} 
                        &      -CLL          &  $d^2_\text{GED}$             &   -CLL     &     $d^2_\text{GED}$      & Dice         &&    -CLL          &    $d^2_\text{GED}$            &     -CLL     &    $d^2_\text{GED}$        &    Dice      \\
                        &&&&&&&$(\times 10^3)$&&$(\times 10^3)$&&\\
                        \midrule
Det.U-Net~\cite{ronneberger2015u}                &    --            &    --           &   --        &    --       &  \textbf{0.727}        &&    --            &  --               &     --      &   --        &  0.624        \\

Prob.U-Net~\cite{kohl2018probabilistic}               &      52.1           &      0.279         &     238.9      &   0.579        &  0.698        &&      4.738          &    0.905          &     4.495      &     0.946       &     0.616     \\

cFlow Net (Planar)               &  \textbf{47.3}               &    \textbf{0.204}            &  \textbf{89.0}         & \textbf{0.288}          &   0.713      &&       \textbf{4.436}         &     \textbf{0.884}          &   \textbf{4.482}       &  0.877      &      0.632    \\
        cFlow Net (Glow) &  {49.2}               &   0.302            &     217.0     &      0.547     &  0.704        &&   4.482             &    0.901          &   4.488        &  0.878         &   0.620    \\  
\bottomrule
\end{tabular}
%\vspace{-0.5cm}
\end{table}

Both variants of the cFlow Net models use $K=4$ flow steps. The \emph{decoder} network in the cFlow Net and Prob. U-Net was a deterministic U-Net with 4 resolutions identical to the ones used in~\cite{kohl2018probabilistic}. Architectures of both \emph{encoder} and \emph{prior} networks were similar to the encoding path of the decoder network. In addition to predicting the mean $\mubf$ and variance $\sigmabf^2$,  the encoder network in the cFlow Net model outputs a context vector $\cbf$ of dimension $H=128$ which is input to the \emph{flow transformation} block as illustrated in Figure~\ref{fig:cFlowNet}. The conditioning network $f_k(\cdot)$ is a three layered multi-layer perceptron (MLP) with 8 hidden units. Latent space dimension of $L=6$ was used for the Prob. U-Net and the cFlow Net models. All the models were trained using a batch size of $96$ and a learning rate of $10^{-4}$ with the Adam optimizer~\cite{kingma2014adam}. The models were trained for a maximum of $300$ epochs and training convergence was assumed when there was no improvement in validation loss for $20$ epochs. Models with the best validation loss was used to evaluate the performance on test set reported in Table~\ref{tab:results}. The experiments were run using PyTorch~\cite{paszke2019pytorch}
%\footnote{Open source implementation of our model will be made available here: \url{https://github.com/raghavian/cFlow}} 
on a single Tesla K80 GPU with 12GB memory. The computation time for both variants of the cFlow Net models on LIDC dataset was $250$s, and about $30$s on the Retinal Vessel dataset per training epoch. The average CO$_2$ footprint of {\emph developing} and training the baseline and proposed models is estimated to be 22.3 kg or equivalently about $180$ km traveled by a car, measured using Carbontracker~\cite{anthony2020carbontracker}. 

\vspace{-0.2cm}
\subsection{Results \& Discussion}
\label{sec:results}
Performance of all the models on test set of both the datasets are reported in Table~\ref{tab:results}. Within each dataset we report the performance when compared to \emph{All Raters} and a \emph{Single Rater}. Statistically significant improvement in performance (based on paired sample t-tests with $p<0.05$) when compared to other models are highlighted in bold.

\begin{figure}[t]
    \centering
    \includegraphics[width=0.89\textwidth]{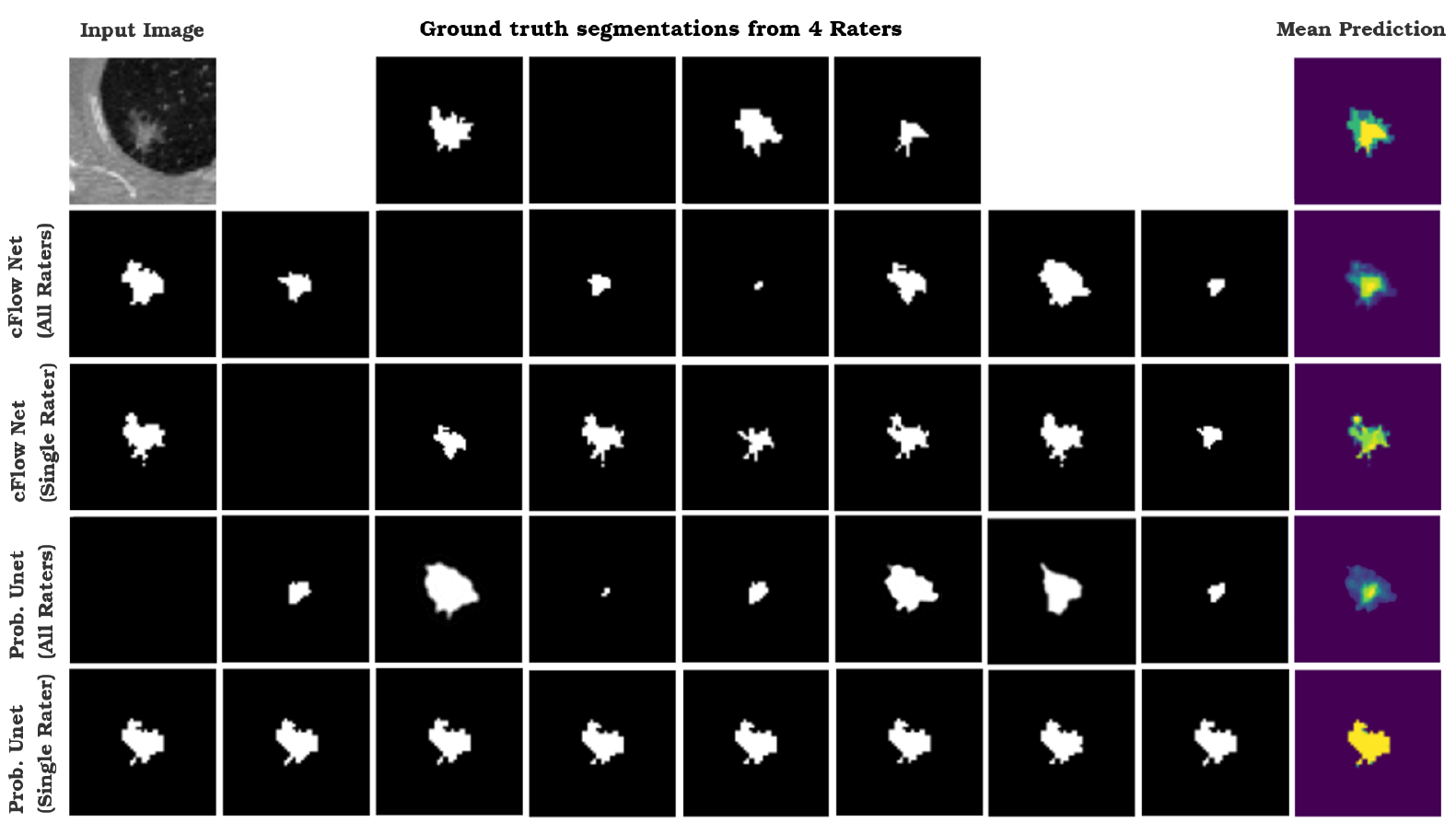}
    \caption{Qualitative results showing the segmentation diversity of the cFlow Net model and Prob. U-Net for one scan from LIDC-IDRI test set. First row shows the input image, segmentation masks from the four raters; Rows 2 and 3 are samples from cFlow Net model when trained with all and a single (first) rater; Rows 4 and 5 show samples from the Prob. U-Net model for all and single rater setting. Mean prediction over all samples are shown in the last column (brigher regions correspond to higher probability).}
    \label{fig:allExperts}
    \vspace{-0.5cm}
\end{figure}

The proposed cFlow Net (Planar) model is consistently better than the baseline Prob. U-Net model on the LIDC dataset in $d^2_\text{GED}$ and -CLL measures. The performance of the cFlow Net (Planar) model in the \emph{Single Rater} setting shows a large improvement when compared to Prob. U-Net model. This is also demonstrated in Figure~\ref{fig:allExperts} seen as more realistic and diverse samples generated only by training only on a single (the first) rater. There is a small reduction in performance of all the conditional generative models when compared to the Det. U-Net model in dice accuracy. 

The significant improvements in $d^2_\text{GED}$ for the cFlow Net models reported in Table~\ref{tab:results} are also reflected qualitatively in the samples shown in Figure~\ref{fig:allExperts}. Samples from cFlow Net (row 2) are not only able to capture the variations amongst all four raters (row 1) but the remainder samples appear plausible. When trained with a single rater (row 3), the cFlow Net model is still able to capture a richer diversity of segmentations. As annotations are available from only a single rater in majority of applications, this behaviour of the cFlow Net of being able to capture diverse segmentations from single rater is desirable. This is in contrast with the samples from Prob. U-Net even when trained with all raters (row 4), where the samples appear blurry and are unable to reflect the diversity of the four raters. This lack of diversity becomes more pronounced when trained with a single rater, as the Prob. U-Net model outputs almost identical looking samples (row 5).

In the last column of Figure~\ref{fig:allExperts} we also show the mean prediction obtained from samples of each model (brighter regions have higher probability). The mean predictions from the cFlow Net model trained on a single rater could be more informative than the mean prediction from Prob. U-Net trained on a single rater. This further strengthens our argument that improving the approximation to the latent posterior distribution with conditional normalizing flows helps capture meaningful uncertainty with the possibility of sampling unlimited number of diverse segmentations.

%This reduction in dice accuracy can be attributed to the use of thresholded (at $p \geq 0.5$) mean segmentation obtained from diverse samples.

A similar trend is also observed with the Retinal Vessel dataset. This is a far more challenging dataset as the images are acquired differently and the quality of annotations vary between the six annotators. This is captured as higher $d^2_\text{GED}$ and -CLL across all models. Even within this setting, the cFlow Net models fare better than the Prob. U-Net model in both the single and multiple rater experiments. There was no significant difference in dice accuracy between any of the methods indicating the stochastic generative components of the proposed models do not affect segmentation accuracy.
%\todo{Discuss results on Retinal dataset}

\vspace{-0.3cm}
\section{Conclusion}
\vspace{-0.2cm}

We proposed a novel conditional generative model based on conditional normalizing flows to quantify uncertainty in segmentations. The use of cFlow steps improved the approximation of the latent posterior distribution, captured in the smaller negative conditional log likelihood values and also manifested in the diversity of samples. The primary contribution in this work is the incorporation of conditional normalizing flows for handling high dimensional data such as medical images. The \emph{flow transformation} block is modular and can be easily replaced with any suitable normalizing flow providing access to a rich class of improved conditional generative models~\cite{papamakarios2019normalizing}. We demonstrated this feature of cFlow Net with two types of normalizing flow transformations: Planar~\cite{rezende2015variational} and Glow~\cite{kingma2018glow} with promising performance. 


\subsubsection*{Acknowledgements} We thank Oswin Krause and the Medical Image Analysis group at DIKU for fruitful discussions and valuable feedback.

\bibliographystyle{splncs04}
\bibliography{M335.bib}

\newpage
\section{Supplementary Material}

\subsection{Derivation of cFlow Net objective}
\label{sec:objective}
We start with the motivation of approximating the latent posterior distribution with a variational distribution and minimizing the \emph{reverse} KL divergence,
\begin{align*}
%\scriptstyle
    &\KL \Big[   q(\zbf|\sbf,\xbf) || p(\zbf|\sbf,\xbf)\Big] = \Em_{q(\zbf|\sbf,\xbf)} \Big{[} \log \frac{q(\zbf|\sbf,\xbf)}{p(\zbf|\sbf,\xbf)} \Big{]} = \Em_{q(\zbf|\sbf,\xbf)} \Big{[} \log \Big(\frac{q(\zbf|\sbf,\xbf)}{p(\sbf|\zbf,\xbf)} \frac{p(\sbf|\xbf)}{p(\zbf|\xbf)} \Big) \Big{]}
		\vspace{-0.3cm}
\end{align*}
		\vspace{-0.3cm}
The last equality is due to Bayes' Rule. A little rearranging yields,
\begin{align*}
 &\KL \Big[   q(\zbf|\sbf,\xbf) || p(\zbf|\sbf,\xbf)\Big]    = \Em_{q(\zbf|\sbf,\xbf)} \Big{[} \log \frac{q(\zbf|\sbf,\xbf)}{p(\zbf|\xbf)} \Big{]} - \Em_{q(\zbf|\sbf,\xbf)} \Big{[} \log {p(\sbf|\zbf,\xbf)} \Big] + \log{p(\sbf|\xbf)} 
    \end{align*}
Note the first term is $\KL \Big[  q(\zbf|\sbf,\xbf) || p(\zbf|\xbf)\Big]$ and the first two terms form a lower bound on the conditional likelihood. Thus,
\begin{equation}
    \KL \Big[  q(\zbf|\sbf,\xbf) || p(\zbf|\sbf,\xbf)\Big] =  - \Lcal b\big(p(\sbf|\xbf)\big) + \log p(\sbf|\xbf)
\end{equation}
where
\begin{equation}
     \Lbcal\big(p(\sbf|\xbf)\big) = -\KL \Big[  q(\zbf|\sbf,\xbf) || p(\zbf|\xbf)\Big] + \Em_{q(\zbf|\sbf,\xbf)} \Big{[} \log {p(\sbf|\zbf,\xbf)} \Big]
\end{equation}
The bound $\Lbcal\big(p(\sbf|\xbf)\big)$ is equal to the conditional likelihood when the KL divergence between the variational distribution and the true posterior is zero. This is the reason we can optimize a surrogate objective such as the bound on the conditional likelihood to indirectly optimize the KL divergence.

The negative of the bound on the conditional log likelihood is the standard objective in a cVAE given in Eq.~\eqref{eq:cVae}. With $K$ steps of normalizing flows, we can factorise the random variable at step $K$ with first order Markov assumption as
\begin{equation}
\log q(\zbf_K|\sbf,\xbf)    = \log \prod_{k=1}^{K} q(\zbf_{k}|\zbf_{k-1},\sbf,\xbf)
\end{equation}
The transformed densities at each step are related by the flow transformation in Eq.~\eqref{eq:flowTrans}
\begin{equation}
    \log q(\zbf_k|\sbf,\xbf) = \log q(\zbf_{k-1}|\sbf,\xbf) + \log \Big{|} J_{T_k}(\zbf_{k-1}|\xbf)\Big{|}.
\end{equation}
Following this, the full factorisation with $\zbf=\zbf_K$ can be written as:
\begin{equation}
\log q(\zbf|\sbf,\xbf) = \log q(\zbf_0|\sbf,\xbf) - \sum_{k=1}^K  \log \Big{|} J_{T_k}(\zbf_{k-1}|\xbf)\Big{|}, 
\label{eq:flowSup}
\end{equation}
where we set the transformed variable at step $K$ to be the posterior variable of interest i.e., $\zbf=\zbf_K$. Using this flow transformed distribution from Eq.~\eqref{eq:flowSup} in Eq.~\eqref{eq:cVae} yields the final cFlow Net objective.
\begin{align}
    \mathcal{L}_{\text{cFlow}} 
& = -\Em_{q(\zbf_0|\sbf,\xbf)} \Big{[} \log p(\sbf|\zbf,\xbf)\Big]  \nonumber \\
     &+ \KL \Big[q(\zbf_0|\sbf,\xbf) || p(\zbf|\xbf)\Big] - \Em_{q(\zbf_0|\sbf,\xbf)} \Big{[} \sum_{k=1}^K  \log \Big{|} J_{T_k}(\zbf_{k-1};\xbf)\Big{|}\Big]
     \label{eq:cFlowSup}
\end{align}

\subsection{cFlow Net (Glow)}
\label{sec:glow}
We also demonstrated the performance of cFlow Net model which uses generative flow (Glow) model~\cite{kingma2018glow} with conditioning to transform the base distribution. We followed a strategy similar to~\cite{lu2019structured} to obtain conditional Glow steps wherein we use a neural network which takes the context vector $\cbf$ as input to predict the parameters of the Glow steps.

Each Glow step comprises three sub-steps and transforms the random variable $\zbf_k$ conditioned on the context vector $\cbf$ into $\zbf_{k+1}$:
%\begin{enumerate}

				\vspace{0.2cm}
				\begin{minipage}[h]{0.45\textwidth}
%    \item 
			Sub-step 1. {\bf ActNorm:}
    \begin{align}
        \sbf_k, \bbf_k & = f_k(\cbf) \\
        \zbf_k^{(1)} &= \sbf_k \odot \zbf_k + \bbf_k \\
			\text{Log Det.} &= \sum \log |\sbf_k| 
	\end{align}
%    \item 
	Sub-step 2. {\bf 1x1 Convolution:}
    \begin{align}
        \Wbf_k & = g_k(\cbf) \\
        \zbf_k^{(2)} &= \Wbf_k \zbf_k^{(1)} \\
		\text{Log Det.} &= \sum\log |\text{det}\Wbf_k|
    \end{align}
				\end{minipage}
    %\item 
				\hspace{0.5cm}
				\begin{minipage}[h]{0.45\textwidth}

				\vspace{0.5cm}
	Sub-step 3. {\bf Affine Coupling:}
    \begin{align}
    \zbf_{a,k}^{(2)}, \zbf_{b,k}^{(2)} &= \text{split}(\zbf_{k}^{(2)}) \\
    (\log \rbf_k, \tbf_k) &= h_k( \zbf_{b,k}^{(2)}, \cbf) \\
    \zbf_{a,k}^{(3)} &= \rbf_k \odot \zbf_{a,k}^{(2)} + \tbf_k \\
    \zbf_{b,k}^{(3)} &= \zbf_{b,k}^{(2)} \\
    \zbf_{k+1} &= \text{concat}(\zbf_{a,k}^{(3)},\zbf_{b,k}^{(3)}) \\
	\text{Log Det.} &= \sum \log |\rbf_k|
    \end{align}
				\end{minipage}

with $f_k(\cdot)$, $g_k(\cdot)$ and $h_k(\cdot)$ are MLPs, $\zbf_k^{(1)}, \zbf_k^{(2)}, \zbf_k^{(3)}$ are intermediate transformed variables  and $\zbf_{k+1}$ is the transformed variable after one complete Glow step.
%\end{enumerate}
\vspace{-0.3cm}

\subsection{Estimation of conditional log likelihood}
\label{sec:cll}
The the conditional log likelihood reported in Table~\ref{tab:results} is obtained by marginalizing over the latent variable $\zbf$ approximated using a Monte Carlo estimate with $N=128$ samples for each image $\xbf$ in the test set:
\vspace{-0.3cm}
\begin{equation}
		\log p(\sbf|\xbf) = \log \int p(\sbf|\xbf,\zbf) p(\zbf|\xbf) d\zbf \approx \log \frac{1}{N} \sum_{n=1}^{N} p(\sbf|\xbf,\zbf^{(n)}) p(\zbf^{(n)}|\xbf)
\end{equation}

\vspace{-1.0cm}
%\begin{comment}

%\subsection{Retina Vessel Segmentation dataset}
\begin{figure}[h]
    \centering
    \includegraphics[width=0.4\textwidth]{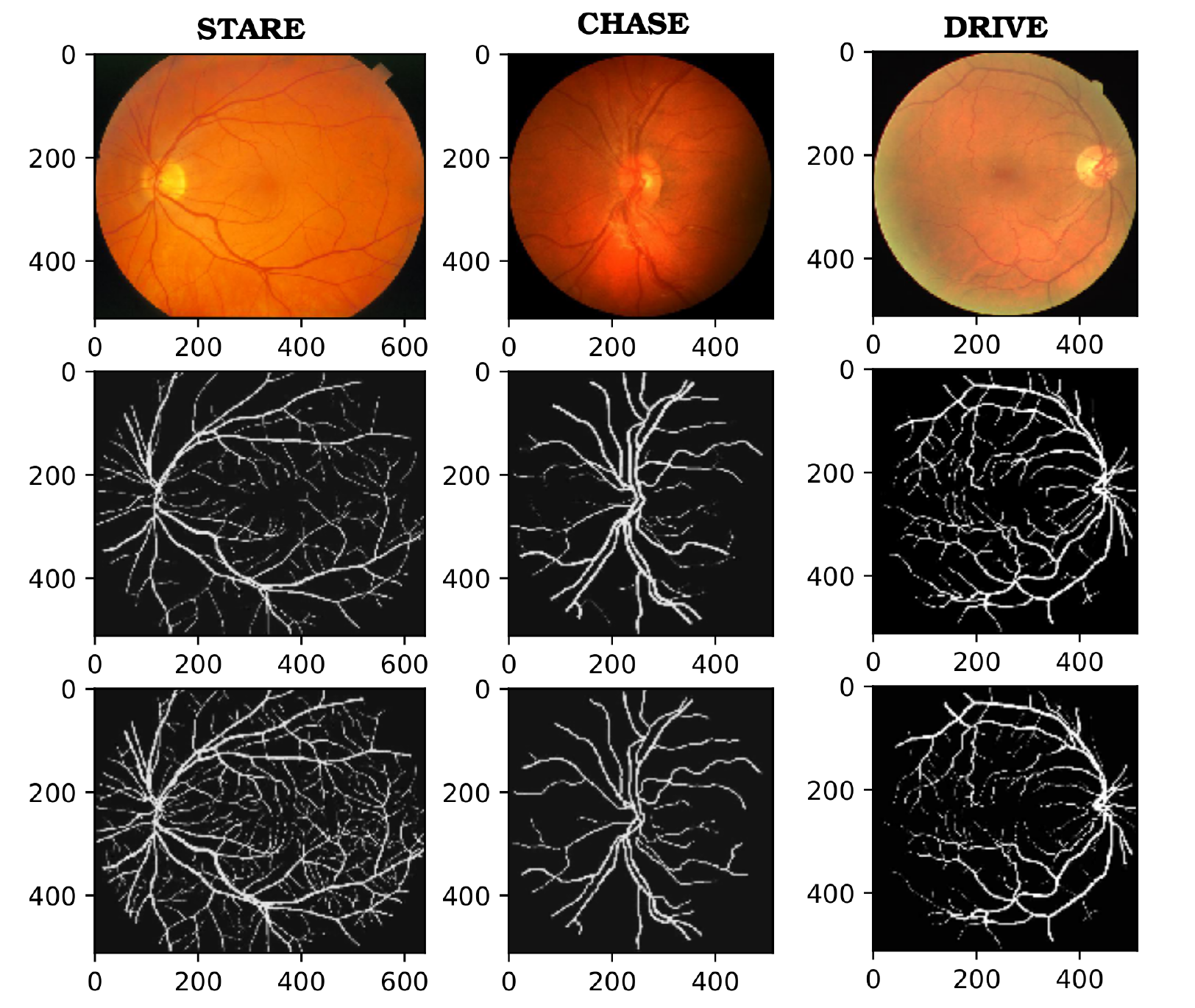}
\vspace{-0.5cm}
		\caption{Examples from the Retinal Vessel Segmentation dataset}
		%. One example from each of the three component databases -- STARE, CHASE, DRIVE -- showing the diversity of images along with two segmentations per image are shown.}
\label{sec:retina}
\end{figure}
%\end{comment}
\end{document}